\documentclass[letterpaper, 10 pt, conference]{ieeeconf}  

\IEEEoverridecommandlockouts                              

\usepackage{cite}
\usepackage{amsmath,amssymb,amsfonts}
\usepackage[linesnumbered,ruled,lined]{algorithm2e}

\usepackage{graphicx}
\usepackage{subcaption}
\usepackage{textcomp}
\usepackage{booktabs}
\usepackage{multirow}
\usepackage{makecell}
\usepackage{tabularx}
\usepackage{xcolor}

\title{\LARGE \bf
LITE: A Learning-Integrated Topological Explorer for Multi-Floor Indoor Environments
}

\author{Junhao Chen$^{1}$, Zhen Zhang$^{1}$, Chengrui Zhu$^{1}$, Xiaojun Hou$^{1}$, 
Tianyang Hu$^{1}$,
Huifeng Wu$^{2}$, Yong Liu$^{1,*}$
\thanks{$*$ Corresponding author. E-mail: yongliu@iipc.zju.edu.cn}
\thanks{$^{1}$ are with the Institute of Cyber-Systems and Control, Zhejiang University, Hangzhou 310027, China.}
\thanks{$^{2}$ is with the Hangzhou Dianzi University, Hangzhou 310018, China.}
\thanks{This work is supposed by Natural Science Foundation U21A20484.}
}

\begin{document}

\maketitle
\thispagestyle{empty}
\pagestyle{empty}

\begin{abstract}

This work focuses on multi-floor indoor exploration, which remains an open area of research. 
Compared to traditional methods, recent learning-based explorers have demonstrated significant potential due to their robust environmental learning and modeling capabilities, but most are restricted to 2D environments. 
In this paper, we proposed a learning-integrated topological explorer, LITE, for multi-floor indoor environments. LITE decomposes the environment into a floor-stair topology, enabling seamless integration of learning or non-learning-based 2D exploration methods for 3D exploration.
As we incrementally build floor-stair topology in exploration using YOLO11-based instance segmentation model, the agent can transition between floors through a finite state machine.
Additionally, we implement an attention-based 2D exploration policy that utilizes an attention mechanism to capture spatial dependencies between different regions, thereby determining the next global goal for more efficient exploration.
Extensive comparison and ablation studies conducted on the HM3D and MP3D datasets demonstrate that our proposed 2D exploration policy significantly outperforms all baseline explorers in terms of exploration efficiency.
Furthermore, experiments in several 3D multi-floor environments indicate that our framework is compatible with various 2D exploration methods, facilitating effective multi-floor indoor exploration.
Finally, we validate our method in the real world with a quadruped robot, highlighting its strong generalization capabilities.

\end{abstract}

\section{INTRODUCTION}

Autonomous exploration is a fundamental problem in the development of 
embodied intelligence and plays a crucial role in uncertain scenarios 
such as search and rescue \cite{niroui2019deep}, 
scene reconstruction \cite{kompis2021informed}, 
and extraterrestrial planetary exploration \cite{tagliabue2020shapeshifter}.
An exploration task requires an autonomous agent to make long-term decisions 
and efficiently construct an environment representation without prior information.
This task remains particularly challenging in indoor environments, 
which are often cluttered and contain multiple floors, 
making autonomous exploration difficult for robots.

In traditional exploration paradigms \cite{yamauchi1997frontier, bircher2016receding, dang2020graph}, autonomous exploration is achieved by iterative online viewpoint sampling and path generation. 
During each iteration, the identified frontiers or sampled viewpoints are evaluated by a cost function to select the most informative one for exploration\cite{gonzalez2002navigation}\cite{kulich2011distance}.
However, viewpoints determined by manually calculated cost functions often lead to locally optimal but short-sighted decisions.
Consequently, the robot may exhibit zigzag movements and leave certain areas unexplored, \textit{e.g.}, narrow corridors or small rooms, rendering the exploration procedure inefficient.

Recently, several learning-based methods, particularly those using deep reinforcement learning, have been proposed to achieve efficient and non-myopic exploration \cite{chaplot2020learning, cao2023ariadne, schmid2022fast}.
These methods learn from extensive interactions with the environment, enabling the trained policy network to extract features from observations and make sequential decisions without exhibiting myopic behavior. 
However, current learning-based approaches are still mostly restricted to 2D environments to simplify exploration modeling, yet they fail to address the challenges of multi-floor indoor environments.



\begin{figure}
    \includegraphics[width=\linewidth]{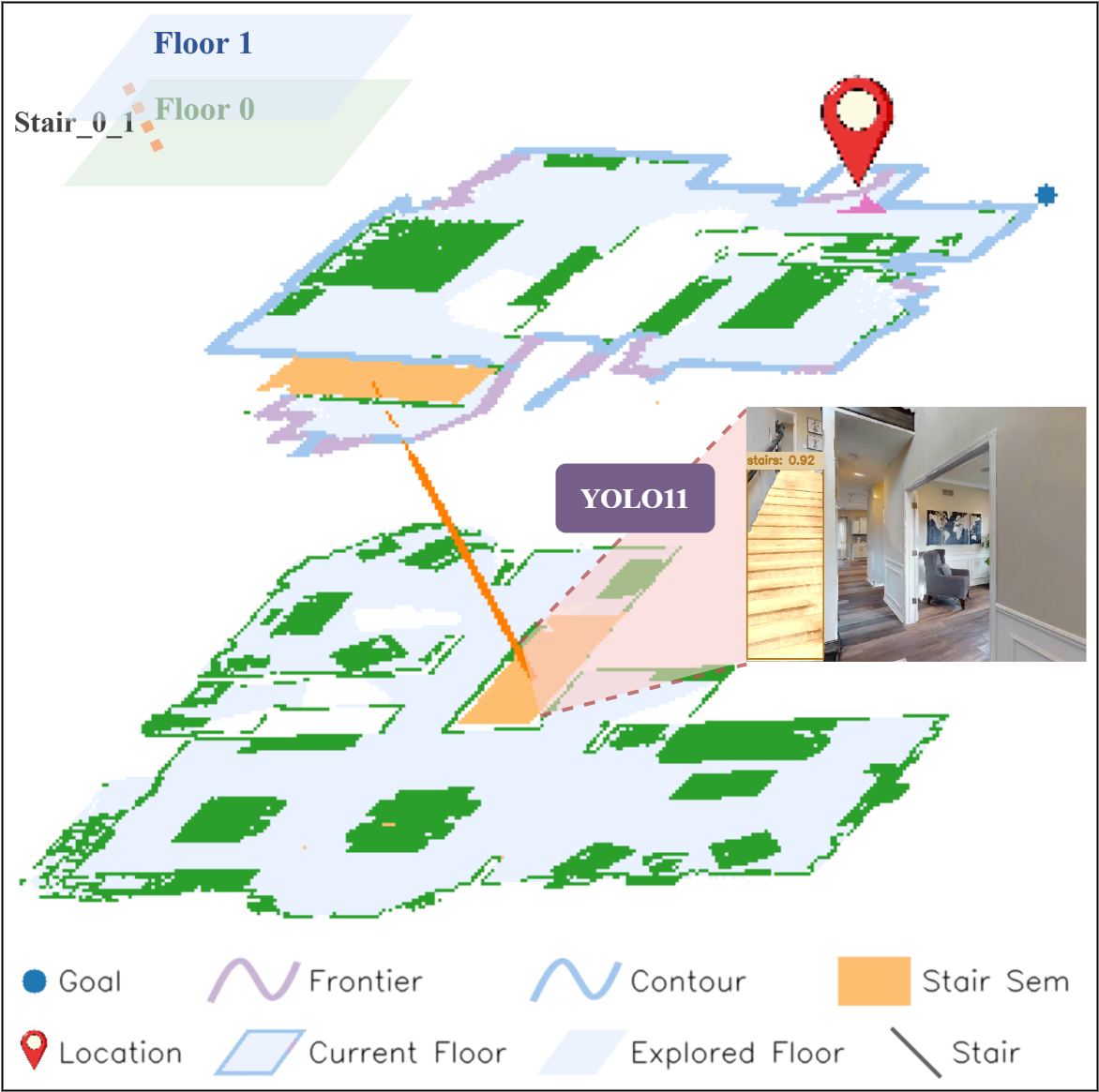}
    \vspace{-10pt}
    \caption{Illustration of LITE. Right: In 
    single-floor exploration, a YOLO11 segmentation model is
    applied to detect stairs.
    Left: The stair information is utilized to build floor-stair 
    topology incrementally, which allows transition between multiple floors.
    Middle: LITE starts exploration from the first floor using 
    an attention-based 2D exploration policy and then guides the agent upstairs
    to achieve multi-floor exploration.}
    \vspace{-15pt}
    \label{fig1}
\end{figure}

In this work, we propose a novel 
\textbf{L}earning-\textbf{I}ntergrated \textbf{T}opological \textbf{E}xplorer,
named \textbf{LITE}, to learn an efficient attention-based 2D exploration policy and integrate it into multi-floor indoor environments (see Fig. \ref{fig1}). 
LITE decomposes the multi-floor environments into a topological graph, where each node represents a floor, and each edge corresponds to a stair connecting two floors. 
Our key insight is that the information in indoor 3D environments is dense within a floor but sparse between floors.
Consequently, a floor-stair topological representation is sufficient and \textit{lite} for exploration and downstream tasks. Furthermore, it can be easily integrated with both learning and non-learning methods.
LITE utilizes a finite state machine (FSM) to manage the exploration process within the floor-stair topology, dividing it into four distinct states, 
while a YOLO11-based\cite{khanam2024yolov11} instance segmentation model is used for real-time stair prediction and topology building.
This design facilitates the integration of various 2D exploration policies for exploring multi-floor indoor environments.
In this work, we first focus on enhancing sustained efficiency of our attention-based 2D exploration policy. 
This policy captures spatial information using a vision transformer, generating near-optimal goals that direct the agent toward high-information regions.
We then integrate this 2D exploration policy into the floor-stair topology, enabling multi-floor indoor exploration without requiring high-dimensional feature inputs.

Our contributions are summarized as follows:
\begin{itemize}
\item We propose a novel method called \textbf{LITE} to explore indoor
multi-floor environments using floor-stair topology. We develop a new
paradigm for seamlessly integrating different 2D exploration methods into 
3D exploration, leveraging the advantages of learning-based approaches.  
\item We design an attention-based 2D exploration policy in LITE
to select global navigation goals effectively. Extensive experiments
demonstrate that our algorithm achieves
state-of-the-art 2D exploration performance.
\item We validate our LITE on a quadruped robot in the real world, showing its strong generalizability of our method. 
\end{itemize}

\section{Related Works}

\subsection{Traditional Exploration methods}
Traditional exploration methods often begin with Simultaneous Localization and Mapping (SLAM), followed by selecting navigation viewpoints for further exploration.
Frontier-based exploration, introduced by Yamauchi in \cite{yamauchi1997frontier}, identifies frontiers (\textit{i.e.}, the points between explored and unexplored areas) in the partially built map and greedily selects the closest one as the next goal. 
Improvements include cost functions that balance gain and distance \cite{gonzalez2002navigation, kulich2011distance} or multi-resolution maps \cite{batinovic2021multi} to improve exploration efficiency. 
In contrast, sampling-based methods \cite{bircher2016receding, witting2018history, schmid2020efficient} maintain a Rapid-Exploration Random Tree (RRT) to sample stochastic viewpoints and select the best viewpoint with the highest gain.
However, the process is usually computationally intensive due to extensive samplings and gain computation, and has a limited capacity for environmental comprehension.

\subsection{Learning-based methods}
Recently, deep learning has become a promising approach in autonomous robot exploration.
While frontier-based and sampling-based methods are inherently short-sighted, deep learning methods enable the modeling of complex patterns from massive data, leading to more robust and adaptive exploration strategies. 
Different learning-based exploration approaches include graph neural networks\cite{chen2020autonomous}, generative models\cite{schmid2022fast}, and imitation learning\cite{reinhart2020learning}, with reinforcement learning being the most prevalent\cite{lodel2022look, sivashangaran2023deep}.
For instance, \cite{niroui2019deep} integrates deep reinforcement learning with frontier-based methods for optimal frontier selection.
Neural-SLAM \cite{chaplot2020learning} utilizes an end-to-end neural network to directly generate an action for the agent, though the training process is challenging due to simultaneous optimization.
ARiADNE \cite{cao2023ariadne} exploits an attention-based network to select goals in a topological graph, which avoids the agent's myopic action.
Subsequent research \cite{cao2024deep} in large-scale environments once again emphasizes the high effectiveness of attention-based methods in capturing long-range spatial dependencies.
However, no one has integrated these approaches into 3D exploration tasks.
Intuitively, extending reinforcement learning observations to the 3D domain introduces excessive complexity, so it is essential to decompose 3D spaces to lower dimensions for effective learning.

\subsection{Exploration in multi-floor indoor environments}
Exploration in multi-floor indoor environments remains challenging, as the robot must consider both horizontal and vertical regions simultaneously.
Approaches designed for 3D spaces like RH-NBV \cite{bircher2016receding}, GBP\cite{dang2020graph}, and A-TARE \cite{kratky2021autonomous} are capable of handling such tasks but must be equipped with aerial robots, which are not suitable for indoor environments.
In the navigation field, \cite{kim2024development} designs a multi-level navigator for robot delivery, which uses a floor topology to guide multi-floor navigation.
We note that the transition between floors does not require a highly accurate 3D map, as they are typically connected only by elevators or stairs.
Consequently, we model the multi-floor indoor environment as a floor-stair topology to leverage the advantages of 2D reinforcement learning algorithms for exploration on each floor. 
To the best of our knowledge, we are the first to integrate reinforcement learning-based exploration algorithms into multi-floor indoor exploration using a topological approach.

\begin{figure*}[htbp]
\centering
\includegraphics[width=\linewidth]{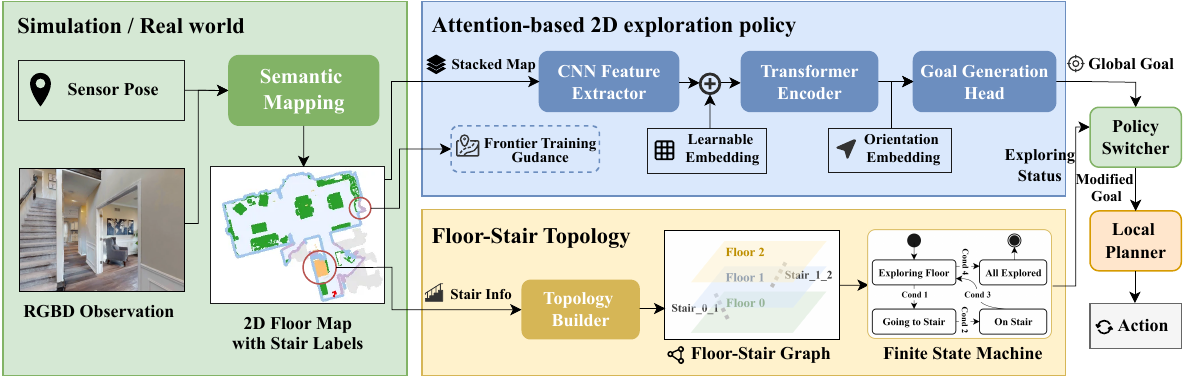}
\caption{The overall architecture of LITE. The semantic mapping module utilizes RGB-D observations to generate the current floor map with stair labels.
Our attention-based 2D exploration policy then extracts long-range features and determines the next informative global goal for single-floor exploration.
Simultaneously, stair information is employed to build a floor-stair topology and select an exploration status within a finite state machine, allowing the agent to transition between multi-floors. 
}
\vspace{-10pt}
\label{architenture}
\end{figure*}

\section{Task Formulation}
Consider an autonomous robot placed in an unknown and bounded 2D environment represented by a grid map $\mathcal{Q} \subset \mathbb{R}^2$. 
The partial map $\mathcal{M} \subset \mathcal{Q} $ is incrementally constructed during exploration using sensor data, comprising free area $\mathcal{M}_f$, 
occupied area $\mathcal{M}_o$ and unexplored area $\mathcal{M}_u$. 
When exploration begins, the covered area is denoted as
$\mathcal{A} = \mathcal{M}_f \cup \mathcal{M}_o$. We follow 
the task setup in \cite{chaplot2020learning}
to define the 2D exploration problem as follows:

\textit{Problem 1}: Given an unknown 2D environment and a time budget $\mathcal{T}$, find the optimal goal selection policy $\pi_{\theta^*}$ to maximize the covered area $\mathcal{A}$ within $\mathcal{T}$:

\vspace{-5pt}
\begin{equation}
\label{eqn_example}
\theta^* = \mathop{\arg\max}_{\theta} \sum_{t = 0}^{\mathcal{T}} \mathcal{A}(a_t | s_t, \pi_{\theta}),
\end{equation}

where $a_t$ and $s_t$ are the action and state at the time $t \leq \mathcal{T}$, $\theta^*$ is the parameters of the optimal policy neural network.
For a multi-floor indoor environment, we expect the robot to cover areas $\mathcal{A}_i$  on all floors given time budget $\mathcal{T}$ using our floor-stair topology and 2D exploration policy.

\textit{Problem 2}: Given an unknown multi-floor indoor environment
and a time budget $\mathcal{T}$, find a floor-stair topology
$\mathcal{G}$ to cover the areas $\mathcal{A}_i$ of all floors using the
optimal 2D exploration policy
$\theta^*$ from \textit{Problem 1}.

\section{Methodology}
In this section, we outline the overall architecture of LITE
and provide a detailed methodology for each module.

\subsection{Overall Architecture}
The overall architecture is illustrated in Fig. \ref{architenture}. 
Firstly, the semantic mapping module utilizes RGB-D observations to create the current floor map, including stair semantic labels generated by YOLO11\cite{khanam2024yolov11} model.
Secondly, our attention-based 2D exploration policy employs a transformer encoder to extract long-range features and determine the next global goal for single-floor exploration. 
The policy leverages frontiers from the floor map and additional embedding information during training to efficiently identify informative areas and minimize redundant exploration.
Concurrently, a floor-stair topology is built incrementally to decompose the multi-floor environment and select an exploration state within a finite state machine, allowing the agent to transition between different floors.
Once the global goal and exploration status are determined, the policy switcher and  local planner module directs the agent's subsequent actions.

\subsection{Map Representation} \label{map}
We represent each floor as a 2D grid map with additional stair semantic information. 
On the one hand, we use a geometric method to generate self-centered point clouds from depth images and camera poses, which are then flattened into a 2D local grid map. 
The grid has two channels, $2 \times L\times L$, representing the obstacle map and the explored map.
By applying pose transformations over time, we merge local maps into a global grid map with dimensions of $2 \times M\times M$. 
Here, $L$ and $M$ denote the width or height of grid map, which contains $L\times L$ or $M\times M$ grid cells at a certain resolution. 
On the other hand, we employ a semantic mapping module similar to \cite{chaplot2020object} to generate an additional stair information map. 
Specifically, we train a fine-tuned YOLO11 model, leveraging self-labeled stair datasets, to perform instance segmentation of stairs and generate corresponding semantic masks.
With the stair masks, the corresponding point clouds are marked as stair spaces and projected into a stair semantic map with the same dimensions of $1 \times M\times M$.

\subsection{Attention-based 2D Exploration Policy}
To achieve efficient exploration on each floor, one significant component of our LITE is an attention-based 2D exploration policy using Proximal Policy Optimization (PPO) algorithm\cite{schulman2017proximal}. 
This policy, denoted as $\pi$, receives map observations and determines the next best goal as an action to maximize exploration rewards.

\textit{1) Observation:} To understand the environment comprehensively, we preprocess the 2D grid map, similar to \cite{chaplot2020learning}, by combining current and historical poses into both local and global maps, expanding their dimensions to $4\times L\times L$ and $4\times G\times G$, respectively. 
We then apply max pooling to the global map to match the local map's size, resulting in a final stacked map of $8\times L\times L$. 
This stacked map $m_t \in [0, 1]^{8\times L \times L}$ serves as the policy's observation $o_t$ at time $t$. 
We avoid using raw RGB-D images as observation since it hinders network convergence 
and exacerbates the sim-to-real gap.

\textit{2) Action:} We define the action of attention-based policy as the global goal $(x, y)$ within the local map $L \times L$ at time $t$:
$
    a_t = (x, y) \sim \pi_\theta(a_t \mid o_t)
$.
Here, $\theta$ represents the policy network's parameters. 
This goal is referred to as a global goal, while a local planner is employed to navigate towards it. 
We utilize the Fast March Method\cite{sethian1996fast} as our local planner to plan a path from the current agent's pose to the goal, determining the final action of the agent.

\textit{3) Policy Network Architecture:}
A significant challenge in exploration is achieving global spatial awareness  and avoiding local optima. 
To address this, we employ a vision transformer\cite{dosovitskiy2020image} paradigm, as illustrated in the right top part of Fig. \ref{architenture}. 
This transformer captures long-range dependencies across the map observation using its self-attention mechanism. 
A CNN feature extractor first processes the stacked map to produce high-level features with $C \times P \times P$ dimensions. 
Then, we split the features into $P^2 \times C$ and combine them with a learnable positional embedding, after which an attention layer learns dependencies between different spatial 
patches. 
This attention-based encoder enables the policy to focus on the most relevant parts of the input, leading to more accurate decision-making.
Finally, we design a goal generation head as a decoder for generating goal action $a_t$ within the local map. 
This head utilizes fully connected layers and a normal distribution to generate the final goal $(x, y)$.
Additionally, we incorporate an orientation embedding for efficient movement during exploration.

Given that we use PPO for training, the goal-generation head is designed with two branches: one predicts the goal action $a_t$, while the other estimates the state-value (\textit{i.e.} $V(s) = \mathbb{E}_{\pi} \left[ \sum_{t=0}^{\infty} \gamma^t r(s_t , a_t) \right]$, where $\gamma$ is the discount factor and $r$ is the reward function). 
Such an architecture is straightforward to train and converge, and it quickly learns informative goals from interactions with the environment.

\textit{4) Rewards with Frontier Guidance:} To address Problem 1, we design a reward function with two components to facilitate efficient exploration: $r_t(o_t, a_t) = r_g + r_f \cdot \lambda$, where $\lambda$ is a weight hyper-parameter for balancing the components. 
The first component $r_g = \mathcal{A}_t - \mathcal{A}_{t-1}$ is area information gain navigating to the global goal $a_t$. 
The second part $r_f$ measures the distance between the global goal and frontier edges $D_{f, g}$:
\begin{equation}
    r_f = 
    \begin{cases}
        0, & D_{f, g} \ge D_{max} \\
        0.05 / D_{f, g}, & D_{min} \leq D_{f, g} < D_{max} \\
        0.005, & D_{f, g} < D_{min} \\
    \end{cases}    
\end{equation}
where $D_{max}$ and $ D_{min}$ are the maximum and minimum thresholds distance between the global goal and frontier edges. 
The frontier reward incorporates the benefits of traditional frontier-based methods, guiding the global goal to be generated near frontiers, which represent potentially unexplored regions.
Consequently, the trained policy is likely to explore the environment more efficiently.

\subsection{Floor-Stair Topology} \label{subsection:floor-stair-topology}

Another key component of LITE is the floor-stair topology, which decomposes a multi-floor indoor environment into a topological graph containing 2D floor-maps and stairs (see Fig. \ref{fig1}, and Fig. \ref{architenture}).
The transition within the floor graph is conducted by a finite state machine, including different states: \{\texttt{ExploringFloor}, \texttt{GoingToStair}, \texttt{OnStair}, \texttt{AllExplored}\}.
This structure allows us to seamlessly integrate the learning-based 2D exploration algorithm into the multi-floor exploration without much effort. 
The overall algorithm, detailed in Algorithm \ref {alg:topology-builder}, puts the agent in a completely unknown environment $\mathcal{M}$ and incrementally builds a floor-stair topology for environment representation. The key components of the algorithm will be introduced in the following paragraphs.

{
\begin{algorithm}[!t]
    \caption{Floor-Stair Topology Builder} \label{alg:topology-builder}
    \KwIn{Unexplored multi-floor environment $\mathcal{M}$}
    \KwOut{Floor-stair topology graph $\mathcal{G}$}
    Initialize the agent pose $x_t$ on the first floor of the environment $\mathcal{M}$, and obtain the initial RGB-D observation $\mathcal{O}_t$\;
    $\mathcal{S} \leftarrow \texttt{ExploringFloor}$; $\mathcal{G} \leftarrow \varnothing$\;
    \While{$\mathcal{S} \textup{ is not } \textup{\texttt{AllExplored}}$}{
        Obtain the stair instances $ins_t \leftarrow \textbf{YOLO11}(\mathcal{O}_t)$\;
        Update stair map $l_t \leftarrow \textbf{SemanticMapping}(ins_t)$\;
        \uIf{$\mathcal{S} \textup{ is \texttt{ExploringFloor} or \texttt{GoingToStair}}$} {
            \uIf{$\textup{Arrive at a new floor }i$} {
                Create a new floor node $v_i$\;
                $\mathcal{G}.\textbf{addVertex}(v_i)$\;
                Store the previous stair information\;
            }
            \uIf{$\textup{There is a new stair area in }l_t$} {
                Create a new stair edge $e_{ij}$ based on $l_t$\; 
                $\mathcal{G}.\textbf{addEdge}(e_{ij})$\;
            }
            Update $V$ and $E$ in the topology graph $\mathcal{G}$\;
        } \ElseIf{$\mathcal{S} \textup{ is \texttt{GoingToStair}}$} {
            Update $E$, store the previous floor information\;
        }
        $\mathcal{S}\leftarrow \textbf{UpdateExplorationStatus}(\mathcal{G}, \mathcal{S}, x_t)$\;
        Select exploration policy $\pi \leftarrow \textbf{ChoosePolicy}(\mathcal{S})$\; Generate exploration goal $a_t^{\prime} \sim \pi$\;
        Navigate one step $\mathcal{O}_t, x_t \leftarrow \mathcal{M}.\textup{step}(a_t^{\prime})$\;

    }
    \Return{$\textup{The floor-stair topology graph }\mathcal{G}$}\;
\end{algorithm}
}


\textit{1) Instance Segmentation Model based on YOLO11:} To achieve fast and accurate stair  detection (as described in Algorithm 1, Line 4), we developed a stair instance segmentation model leveraging YOLO11\cite{khanam2024yolov11}. 
Given the scarcity of labeled stair data in most public datasets, we randomly collected 1,511 stair images from the internet and performed semi-automatic annotation using the open-source labeling software Labelme and its integrated SAM model\cite{kirillov2023segany}.
The annotations were then converted into YOLO format for training and validating, after which the model can be deployed to obtain the segmentation masks of stairs. 
As described in Section B, the semantic mapping approach is employed to project mask information into a stair mask map $l_t$ in $1\times M \times M$ (see Algorithm \ref{alg:topology-builder}, Line 5), which is used for subsequent state transitions and policy switching.

{
    \begin{algorithm}[!t]
        \caption{Exploring Status Transition with FSM}  \label{fsm}
        \KwIn{Floor graph $G_t$, Previous exploration status $S_{t - 1}$, Current pose $x_t$}
        \KwOut{Current exploration status $\mathcal{S}_t$}
      $(V_t, E_t) \leftarrow G_t$\;
      $f_{done} \leftarrow \mathbf{IsCurrFloorDone(}V_t\mathbf{)}$\; 
      $g_{all\_visited} \leftarrow \mathbf{IsAllGraphExplored(}V_t, E_t\mathbf{)}$\;
      $e_{on\_stair} \leftarrow \mathbf{IsOnStair(}E_t, x_t\mathbf{)}$\;
      $S_{t} \leftarrow S_{t- 1}$\;
        \uIf{\textup{$\mathcal{S}_{t - 1} = \texttt{ExploringFloor} \textbf{ and } f_{done}$}}{
            \uIf{$g_{all\_visited}$}{
                $\mathcal{S}_{t} \gets \texttt{AllExplored}$ \hfill \(\triangleright\) Cond 4
            }  
            \uElse{
                $\mathcal{S}_{t} \gets \texttt{GoingToStair}$ \hfill \(\triangleright\) Cond 1
            }  
        }
        \uElseIf{\textup{$\mathcal{S}_{t - 1} = \texttt{GoingToStair} \textbf{ and } e_{on\_stair}$}}{
            $\mathcal{S}_{t} \gets \texttt{OnStair}$ \hfill \(\triangleright\) Cond 2
        }
        \ElseIf{\textup{$S_{t - 1} = \texttt{OnStair} \textbf{ and not } e_{on\_stair}$}}{
            $\mathcal{S}_{t} \gets \texttt{ExploringFloor}$ \hfill \(\triangleright\) Cond 3
        }
        \Return{\textup{Current Exploration Status} $\mathcal{S}_t$}\;
    \end{algorithm}
}

\textit{2) Topology Builder:} In a multi-floor indoor environment, to tackle \textit{Problem 2}, we represent the entire environment as a topological graph $\mathcal{G} = (V, E)$, where $V = \{v_i \mid i = 0, 1, \dots, n\}$ denotes floor maps and $E = \{(v_i, v_j), v_i, v_j \in V\}$ denotes edges connecting all floors. 
When exploring a new floor $i$, we add the 2D grid map $v_i$ of size $2 \times M\times M$ to $V$ and continuously update the floor map in exploration(
Line 7-10, 14). 
The trained YOLO11 instance segmentation module is utilized for real-time stair detection and segmentation.
Once we detect a new stair, the topology builder will extend a new edge $v_i, v_{i+1}$ with stair information to $V$(
Line 11-13). 
If a floor is explored or a stair is visited, their attributes in $\mathcal{G}$ (\textit{e.g.} floor obstacle maps, stair semantics in two floors) are updated and stored for future exploration.

\textit{3) Exploration Status Transition:} After updating topological graph $G_t$, a finite state machine (FSM) manages the transition between different floors (see Fig. \ref{architenture} right bottom part), as shown in Algorithm \ref{alg:topology-builder}, Line 18, $\textbf{UpdateExplorationStatus}(\mathcal{G}, \mathcal{S}, x_t)$.
The transition conditions, Cond $1\sim4$, are described in Algorithm \ref{fsm}. 
We firstly extract floor nodes $V_t$ and stair edges $E_t$ from graph $G_t$ and calculate three boolean variables $f_{done}$, $g_{all\_visited}$ and $e_{on\_stair}$.
$f_{done}$ is true when the current exploring floor is fully explored (i.e., explore ratio $> 95\%$ in simulation or no new areas in last $T_{thre}$ time steps) or time budget is reached; $g_{all\_visited}$ is true after all floors $V_t$ and stairs $E_t$ are visited and explored; $e_{on\_stair}$ is true if current pose $x_t$ is within the range of a stair semantic map. 

The exploration starts on the first floor with status \texttt{ExploringFloor}.
When the current floor is done, the status transitions to \texttt{GoingToStair}, directing the agent to the nearest stair area. 
Upon reaching this stair, the status changes to \texttt{OnStair}, guiding the agent upstairs.
After the stair is visited, the status reverts to \texttt{ExploringFloor} 
for further exploration. 
Once all floors are explored and no new stairs are found, the exploration procedure 
concludes with status \texttt{AllExplored}.

\textit{4) Policy Switcher}: To handle diverse scenarios effectively, the agent must select an appropriate policy $\pi$ to determine the action $a_{t}'$ given different exploration status $\mathcal{S}$, as described in Algorithm \ref{alg:topology-builder}, Line 19, \textbf{ChoosePolicy}($\mathcal{S}$).
When the agent is exploring a specific floor ($\mathcal{S} = \text{ExploringFloor}$), our attention-based 2D exploration policy $\pi_{\theta}$ is employed, facilitating efficient single-floor exploration. 
During the process of navigating towards the stairs ($\mathcal{S} = \text{GoingToStair}$), the center point $c_k$ of each detected stair $e_k$ on the current floor are calculated. Subsequently, the center point closest to the agent's current pose $x_t$ is chosen as the goal point $a_t = c_{k^*}$: 
\begin{equation} \label{eq:ch4-going-to-stair-policy}
    \begin{aligned} 
        c_k = &\frac{1}{N}\left(\sum_{(x_i, y_i) \in l_k} x_i, \sum_{(x_i, y_i) \in l_k} y_i\right), N = \sum_{(x_i, y_i) \in l_k} 1 \\
        k^* = &\underset{e_k \in l_t}{\arg\min} \sqrt{(x_t[0] - c_k[0])^2 + (x_t[1] - c_k[1])^2} \\
    \end{aligned}
\end{equation}
where $l_k$ means the area of stair $e_k$ on the stair map $l_t$, $(x_i, y_i)$ denotes the corresponding coordinates in the area of $l_k$, while $x_t$ means agent pose. This approach streamlines the navigation process, allowing the agent to reach the stairs in the most efficient way. 
For $\mathcal{S} = \text{OnStair}$, since a stair typically has a ladder structure, it is projected as a rectangular area in the top-down map. In this map, the short side of the rectangle represents the entrance and exit of the stairs. To guide the agent upstairs, we compute the perpendicular bisector of the short side, and the end of the extension line of the center line on the side away from the entrance is selected as the goal. Once all stairs and floors have been explored ($\mathcal{S} = \text{AllExplored}$), the policy stops exploration.

{
\begin{table*}[htbp]
\centering
\caption{Results of Comparative Study for LITE-2D.}
\label{table1}
\footnotesize
\begin{tabularx}{0.95\textwidth}{c *{8}{>{\centering\arraybackslash}X}} 
\toprule
\multirow{3}{*}{\makecell[c]{Method}} 
    & \multicolumn{4}{c}{HM3D} 
    & \multicolumn{4}{c}{MP3D Generalization} \\
\cmidrule(lr){2-5} \cmidrule(lr){6-9}
    & CR$\uparrow$ & CA($m^2$)$\uparrow$ 
    & APL($m$)$\uparrow$ & SR$\uparrow$ 
    & CR$\uparrow$ & CA($m^2$)$\uparrow$ 
    & APL($m$)$\uparrow$ & SR$\uparrow$   \\
\midrule
Nearest Frontier 
    & 0.835($\pm$0.125) & 62.3($\pm$10.6) 
    & 0.832($\pm$0.212) & 0.230
    & 0.684($\pm$0.264) & 71.4($\pm$33.8) 
    & 1.032($\pm$0.412) & 0.238 \\
Utility Frontier 
    & 0.839($\pm$0.108) & 62.9($\pm$11.6) 
    & 0.752($\pm$0.189) & 0.143
    & 0.718($\pm$0.251)  & 79.3($\pm$43.6)
    & 1.075($\pm$0.482) & 0.265 \\
RH-NBV 
    & 0.896($\pm$0.103) & 67.3($\pm$12.4)
    & 1.032($\pm$0.226) & 0.403
    & 0.745($\pm$0.244) & 84.8($\pm$51.7) 
    & 1.355($\pm$0.593) & 0.321 \\ 
ANS Global\cite{chaplot2020learning}
    & 0.901($\pm$0.097) & 67.5($\pm$11.6) 
    & 1.039($\pm$0.236) & 0.407
    & 0.748($\pm$0.228) & 88.3($\pm$56.7)
    & 1.355($\pm$0.574) & 0.301 \\
\textbf{LITE-2D (Ours)} 
    & \textbf{0.918}($\pm$0.071) & \textbf{69.1}($\pm$11.5)
    & \textbf{1.054}($\pm$0.248) & \textbf{0.473} 
    & \textbf{0.765}($\pm$0.216) & \textbf{90.5}($\pm$56.1)
    & \textbf{1.364}($\pm$0.544) & \textbf{0.323} \\                                
\bottomrule
\vspace{-15pt}
\end{tabularx}
\end{table*}
}

\section{Experiments}
\subsection{Datasets}
We conduct both 2D and 3D exploration experiments on visually realistic Habitat Simulator\cite{szot2021habitat} and HM3D dataset\cite{ramakrishnan2021hm3d} to achieve strong policy generalization capabilities. 
HM3d includes 800 training and 100 validation scenes, most of which are multi-floor scenes. 
For 2D exploration, we filtered out 57 training and 12 validation scenes, which are single-floor and well-reconstructed. We also generalize our trained policy to MP3D\cite{Matterport3D} validation datasets, including 11 large indoor scenes. 
For multi-floor exploration, we select 6 scenes in HM3D with semantic annotations that are navigable between floors.

\subsection{2D Exploration Results} \label{2Dexp}
\textit{1) Training details:} We use the Habitat Simulator for training and evaluation, 
with $480 \times 640$ RGB-D observations. The local and global grid map sizes
are $L = 240$ and $M = 480$, respectively, and the map resolution 
is 0.05m for each grid. The CNN feature is reshaped to $P^2=225$ patches. During training,
each episode has 1000 steps, and if the explored ratio is larger than $95\%$,
the episode is marked success, which will trigger a new episode. 
Each global step is 25 for generating a new goal.
We train our policy on 2 NVIDIA A4000 16 GB GPU for over 15 million frames.

\textit{2) Metrics:} We use four metrics to compare all methods: Coverage Ratio (CR), Coverage Area (CA), Area weighted by Path Length (APL), and Success Rate (SR).
CR represents the ratio of the explored area to the total explorable area at the end of an episode, and CA is the corresponding explored area.
APL is defined as $\frac{CA}{L}$, where $L$ is the episode path length. 
SR is defined as $S_i$, where $S_i$ equals 1 if the coverage ratio exceeds 95\%. 
All metrics are reported as the mean and standard deviation (except for SR, whose standard deviation is not applicable) across all episodes and scenes.

\textit{3) Comparison Analysis:} We compare our attention-based 2D policy, namely LITE-2D, with several learning and non-learning baselines, including Nearest Frontier\cite{yamauchi1997frontier}, Utility Frontier\cite{kulich2011distance}, RH-NBVP\cite{bircher2016receding}, and ANS global\cite{chaplot2020learning}. 
Nearest Frontier chooses the nearest frontier as the next goal, while Utility Frontier selects the frontier with the highest score: $S(f) = U(f) \cdot e^{-\lambda C(f)}$, where $U(f)$ is utility score of frontier $f$, $C(f)$ is the shortest path 
between the agent and the frontier. 
RH-NBVP instead maintains a Rapid-exploring Random Tree (RRT) and calculates a similar score to select the next best viewpoint (NBV) from the tree iteratively.
ANS employs an end-to-end architecture for exploration, including Neural-SLAM, Global Policy, and Local Policy. 
We only use ANS Global policy with the same input and output for a fair comparison.
Note that we transfer frontier-based methods and RH-NBV to the same framework of LITE-2D, and the best node of RH-NBV serves as the global goal.

\begin{figure}[t]
    \centering
    \includegraphics[width=\linewidth]{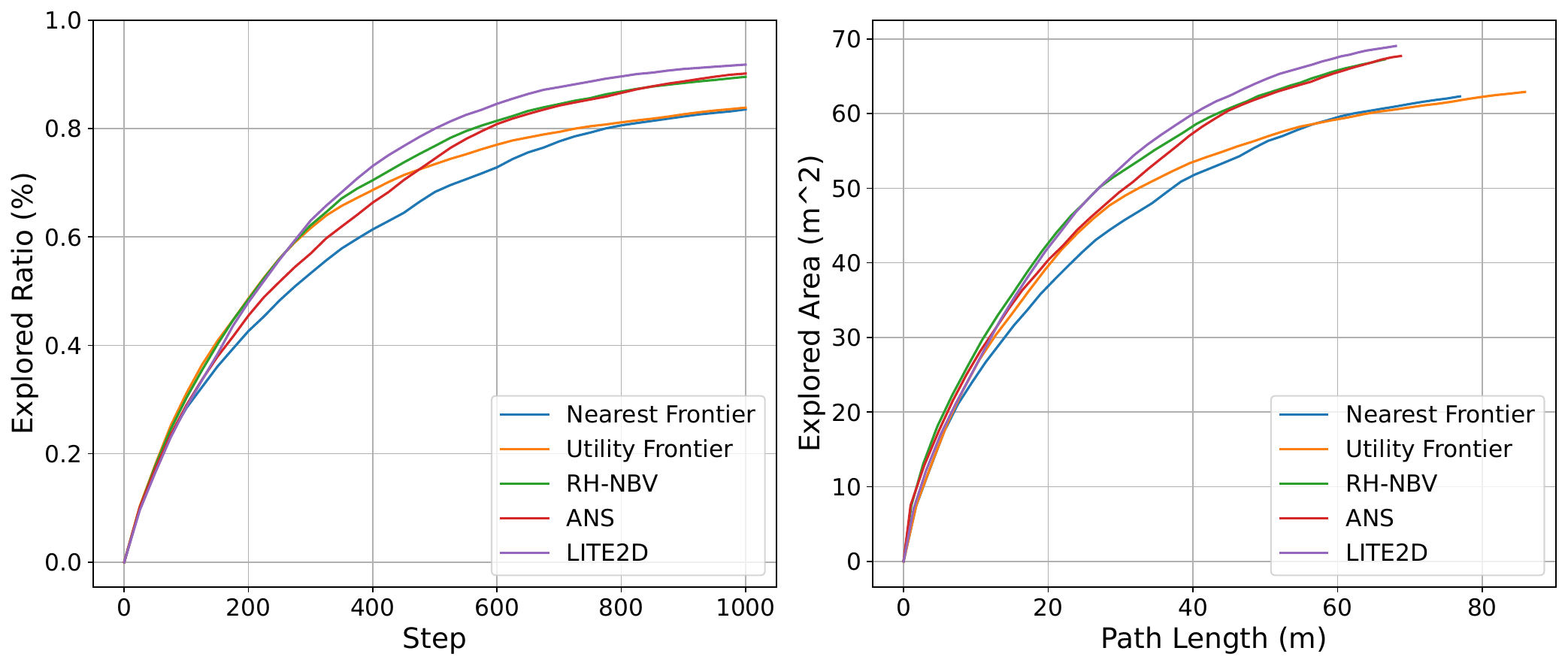}
    \caption{The plot of CR and CA when exploration proceeds.}
    \vspace{-10pt}
    \label{curve}
\end{figure}

The comparison results are reported in TABLE \ref{table1}. The table shows that frontier-based methods perform the worst as they always fall into local optima, leaving some regions unexplored. 
In contrast, sampling-based RH-NBV is more likely to acquire potential information(+6\% CR in HM3D) as long as each regions are sampled.
We utilize frontier edges instead of clustering them to avoid local optima while easily getting potential information from the map.
Compared to the learning-based ANS Global, we achieve significantly better performance (e.g., +1.7\% CR, +1.6 CA, +0.015 APL, and +6.6\% SR).
While ANS Global uses only CNN and fully connected layers as a global policy, our attention-based exploration policy can learn long-range spatial dependencies between different regions and determine a non-shortsighted goal.
Generalization tests on MP3D val with larger scenes show that LITE-2D outperforms
all baselines regarding exploration efficiency. We remark that the SR is much lower 
than expected since there are small reconstruction errors in most scenes and limitations
on episode steps. In most scenes, LITE-2D can 
achieve a coverage ratio higher than 90\% in 1000 steps. Fig. \ref{curve} shows the
average exploration ratio and area when exploration in the exploration process,
where LITE-2D achieves a more efficient exploration
while other baselines experience unnecessary movements.

\begin{figure*}[t]
    \centering
    \includegraphics[width=0.89\linewidth]{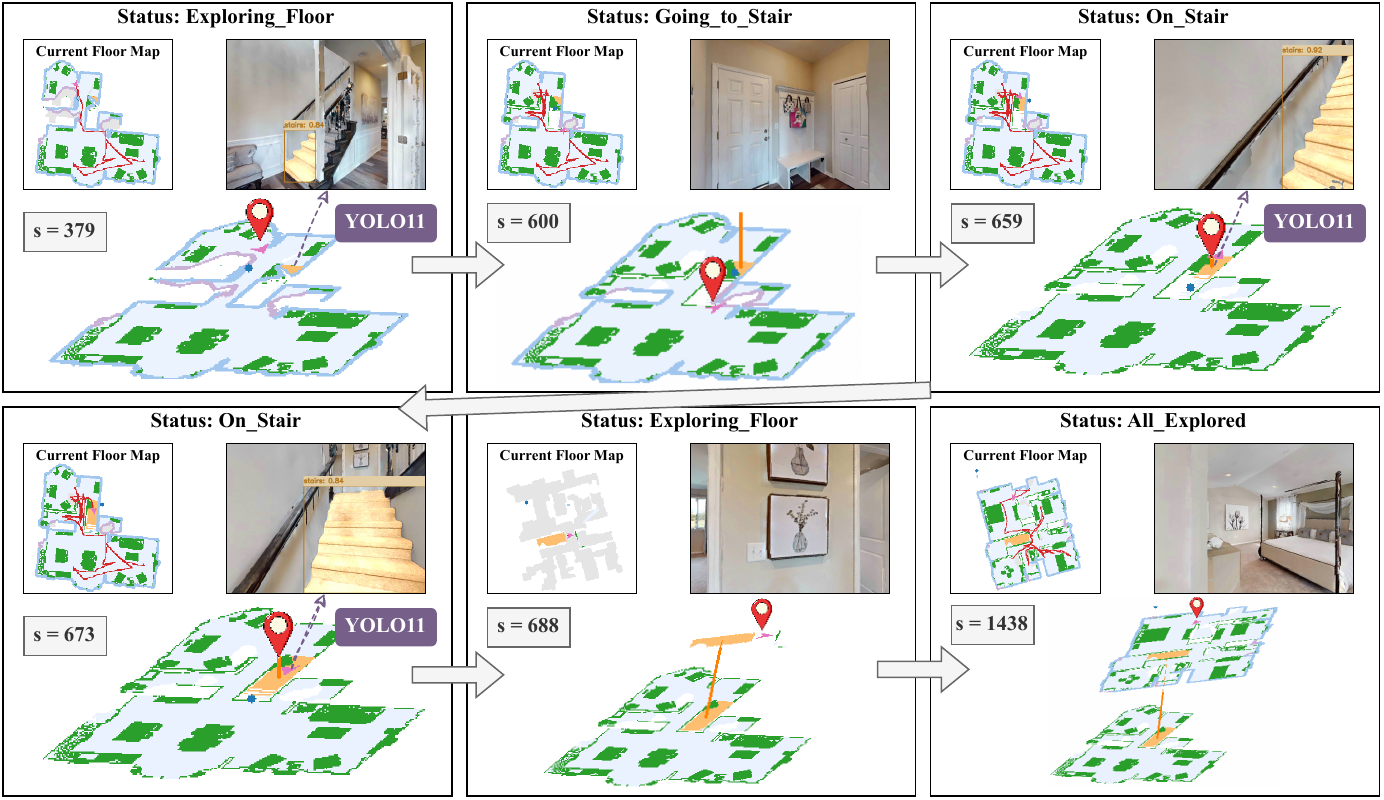}

    \caption{Qualitative result of LITE for multi-floor indoor exploration in HM3D. 
    All visualizations are the same as Fig. \ref{fig1}.}
    \label{vis}
    \vspace{-10pt}
\end{figure*}

\textit{4) Ablation Study:} We consider 3 different variants on HM3D to
validate the effectiveness of each section. 
\begin{itemize}
    \item \textbf{LITE-2D w.o. Attention:} We remove the attention mechanism 
    and directly flatten the features extracted from CNN to select
    a global goal.
    \item \textbf{LITE-2D w.o. Frontier:} We discard the frontier 
    reward from training, with only information gain left. 
    \item \textbf{LITE-2D w.o. Orientation:} The orientation embedding
    is removed in the goal generation head, and the policy has no information 
    about the agent's orientation.
\end{itemize}

The importance of each part in LITE-2D is shown in TABLE \ref{table2}.
Once the attention mechanism is removed, the metrics deteriorate to be similar to ANS Global, indicating that the model fails to focus on key areas effectively. 
In particular, we notice that the APL and SR of LITE-2D w.o. Frontier is much smaller than LITE-2D, which means the last few regions are ignored.
We believe this guidance incorporates additional gain information to the policy so it can learn the underlying unexplored regions of the map. 
Finally, the LITE-2D w.o. Orientation also performs worse than LITE-2D, indicating that incorrect directions can cause the agent to make more turns when navigating to the goal.

\begin{table}[htbp]
\centering
\caption{Results of Ablation Study.}
\label{table2}
\begin{tabularx}{0.95\columnwidth}{c *{4}{>{\centering\arraybackslash}X}} 
\toprule
Method
    & CR$\uparrow$ & CA($m^2$)$\uparrow$ 
    & APL($m$)$\uparrow$ & SR$\uparrow$  \\
\midrule
LITE-2D w.o. Attention
    & 0.902 & 67.7
    & 1.015 & 0.380 \\
LITE-2D w.o. Frontier
    & 0.909 & 68.2
    & 1.007 & 0.373 \\
LITE-2D w.o. Orientation
    & 0.907 & 68.1
    & 1.021 & 0.410 \\
\textbf{LITE-2D} 
    & \textbf{0.918} & \textbf{69.1} 
    & \textbf{1.054} & \textbf{0.473} \\
\bottomrule
\vspace{-15pt}
\end{tabularx}
\end{table}

\subsection{Multi-floor Indoor Exploration Results}
We further perform experiments in 6 multi-floor indoor environments of HM3D to c the ability of floor-stair topology to integrate learning-based 2D exploration policy. 
As stated in Section \ref{subsection:floor-stair-topology}, YOLO11 is employed for real-time instance segmentation of stairs. 
We trained a high-precision model (up to 0.954 $\text{mAP}^{50-95}$) on our stair dataset using the pre-trained YOLO11n-seg model\cite{khanam2024yolov11}, and directly deploy it in our experiments.
We always put the agent on the first floor and explore the environment with floor-stair topology. 
On each floor, we integrate our trained 2D exploration policy for efficient exploration. 

The exploration process of one scene with two floors is shown in Fig. \ref{vis}. 
In this scene, LITE detects a new stair when exploring the first floor and efficiently explores the first floor in 600 steps, reaching an explored ratio of 95\%. 
Then, it extends floor-stair topology and guides the agent upstairs.
No new stairs were detected on the second floor, so the entire exploration process concludes when the second floor is done at 1438 steps.
This demonstrates that our framework effectively integrates 2D learning-based exploration methods in multi-floor environments by utilizing simple floor topology relationships. 
We also integrate all the explorers mentioned in Section \ref{2Dexp} to our floor-topology, comparing the average CR of each floor and total steps when all floors are explored. 
The experimental results in Table \ref{table3} indicate that using our topological explorer, different methods can be applied to explore multi-floor indoor environments, and LITE-2D outperforms other explorers in terms of exploration efficiency. 

\begin{table}[htbp]
\centering
\caption{Results of Multi-floor Exploration with LITE.}
\label{table3}
\setlength{\tabcolsep}{4pt} 
\begin{tabularx}{0.95\columnwidth}{c *{5}{>{\centering\arraybackslash}X}} 
\toprule
Method            & Nearest & Utility & RH-NBV & ANS & \textbf{LITE} \\
\midrule
CR$\uparrow$  & 0.864   & 0.871   & 0.905      & 0.924   & \textbf{0.929}       \\
Steps$\downarrow$ & 1884    & 1767    & 1736       & 1777    & \textbf{1708}       \\
\bottomrule
\end{tabularx}
\end{table}

    

\subsection{Experimental Validation}
\begin{figure}[!t]
    \centering
    \includegraphics[width=\linewidth]{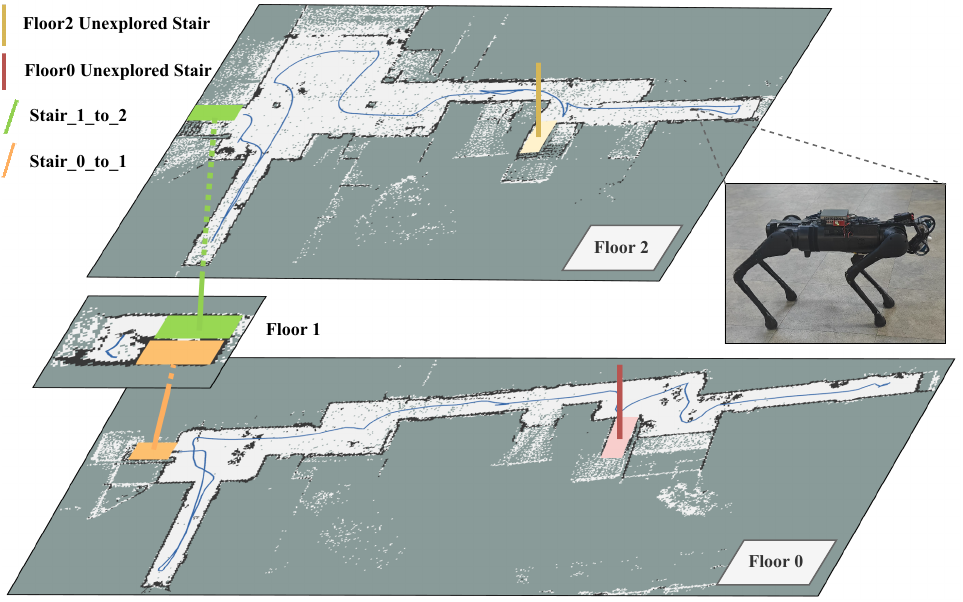} 
    \caption{Final floor-stair topology in the real world}
    \label{validation2}
    \vspace{-10pt}
\end{figure}

We validate LITE in the real world with a quadruped robot, Unitree Aliengo (see Fig. \ref{validation2}). 
To achieve multi-floor exploration, we select a teaching building with four floors, and set the exploration goal to explore the first two floors, and stop the exploration after the second floor has been explored.
Our robot is equipped with a RealSense D435i depth camera and two Livox lidars(for pose estimation). 
Fine-tuned YOLO11-based segmentation model is converted into ONNX format without extra training to predict stairs. 
We use \cite{Zhang2021EfficientMP} as the local planner and learning-based quadrupedal controller for going upstairs. 
The global goal outputted by LITE is re-planned once the goal is reached or the local planner fails. 
On each floor, we maintain a grid map with stair masks, and all floors and stairs are represented as a floor-stair topology (see Fig. \ref{validation2}).
By utilizing an attention-based exploration policy and the floor-stair topology, our algorithm efficiently explores the entire environment despite significant differences in the environment, sensors, and robot models compared to the training scenarios. 
LITE ingeniously constructs the two-floor environment to a three-floor topology, this highlights the strong generalization capabilities of our approach.

\section{Conclusion}
In this work, we propose the first leaning-integrated topological explorer (LITE) for multi-floor indoor exploration. 
LITE contains two key components: an attention-based 2D exploration policy and a floor-stair topology for multi-floor exploration. 
The attention-based 2D exploration policy leverages transformer encoders to capture long-range dependencies of spatial regions and use frontier training guidance to explore unknown areas efficiently. 
With a floor-stair topology, different 2D exploration policies can be seamlessly integrated into multi-floor exploration.
Extensive experiments on both simulation and the real world show our approach's effectiveness and generalization ability. 
Future work will consider more complex topologies for sophisticated multi-floor environments.


\vspace{-5pt}
\bibliographystyle{IEEEtran}
\bibliography{IEEEabrv, ref/reference.bib}

\end{document}